\documentclass[a4paper,11pt]{article}

\usepackage{booktabs}

\usepackage{lineno}
\usepackage[colorlinks=false]{hyperref}
\usepackage{bm}
\usepackage{graphicx}
\usepackage{amssymb}
\usepackage{amsmath}
\usepackage{tabularx}
\usepackage{natbib}
\usepackage{floatrow}
\usepackage{caption}
\usepackage{multirow}
\usepackage{amsmath}
\usepackage{amsfonts}
\usepackage{amssymb}
\usepackage{xcolor}
\usepackage{subfig}
\usepackage{graphicx,amsmath,bm}
\usepackage[binary-units=true]{siunitx}
\usepackage{array}
\usepackage{authblk}
\usepackage{rotating}
\usepackage{enumerate}
\usepackage{ragged2e}
\usepackage{algorithm}
\usepackage{algorithmic}

\usepackage[left=15mm,right=15mm,top=1.5cm,bottom=1.5cm,includeheadfoot]{geometry}
\setcitestyle{square,numbers}
\begin{document}
	
	\title{Contrast-Source-Based Physics-Driven Neural Network for Inverse Scattering Problems}
	
	\author[1]{Yutong~Du}
	\author[1]{Zicheng~Liu}
	
	\affil[1]{\scriptsize Department of Electronic Engineering, School of Electronics and Information, Northwestern Polytechnical University, Xi'an 710029, China}
	\maketitle
	
	\abstract{
		Deep neural networks (DNNs) have recently been applied to inverse scattering problems (ISPs) due to their strong nonlinear mapping capabilities. However, supervised DNN solvers require large-scale datasets, which limits their generalization in practical applications. Untrained neural networks (UNNs) address this issue by updating weights from measured electric fields and prior physical knowledge, but existing UNN solvers suffer from long inference time. To overcome these limitations, this paper proposes a contrast-source-based physics-driven neural network (CSPDNN), which predicts the induced current distribution to improve efficiency and incorporates an adaptive total variation loss for robust reconstruction under varying contrast and noise conditions. The improved imaging performance is validated through comprehensive numerical simulations and experimental data.}
	
	\section{Introduction}
	Electromagnetic inverse scattering imaging \cite{chen2018computationalEMIS} is a technique that reconstructs the material and structure of targets from the measured scattered fields. The applications of inverse scattering imaging include non-destructive testing\cite{Zhou2023PALS}, airport security screening\cite{Ahmed2021MicroSec}, and subsurface inspection\cite{An2024CompoMat}. The challenges of the inverse problems are the nonlinearity caused by multiple scattering effects, and the ill-posedness caused by limited measurement data and propagation loss. The design target of inversion algorithms is to enable accurate, stable, and fast imaging.
	
	Due to the lack of large and diverse real measurement datasets, data-driven deep learning methods\cite{wei2018BPS,Li2018DeepNIS,Liu2022PhaGuideNN,Liu2022SOMnet,ma2023inverse,Du2025QuaDNN} inherently suffer from limited generalization in practical applications. This motivates the use of untrained neural networks in ISPs \cite{Song2022uSOM,Huang2025MatauSOM,Du2025PDNN} with high flexibility and no need for training data. Song et al.\cite{Song2022uSOM} proposed a uSOM-Net to alternately update the induced current and contrast with physical losses defined from both the data and the state equations. Du et al.\cite{Du2025PDNN} proposed a physics-driven neural network (PDNN) solver to leverage the governing physical laws to guide the updating of network hyperparameters to ensure physics-consistent solutions, and apply morphological processing to identify subregions containing true scatterers, thereby reducing the computational burden. However, uSOM cannot accurately reconstructs targets with complex structures. PDNN updates the relative permittivity iteratively, requiring scattered field computations at each step. Even with sub-region selection, the computational speed is still not fast enough. Both solvers fall short of practical application requirements.
	
	In this paper, the contrast-source-based physics-driven neural network (CSPDNN) is proposed to solve ISPs. Instead of directly estimating the relative permittivity, the neural network predicts the induced current distribution, thereby avoiding the explicit matrix inversion of PDNN and significantly improving computational efficiency. Furthermore, an adaptive weight updating strategy is introduced for the total variation (TV) regularization term in the loss function, enabling the solver to automatically adjust to different contrast levels and noise conditions. The effectiveness and robustness of the proposed solver are demonstrated through comprehensive numerical simulations and experimental validations.
	
	\section{Formulation of ISPs}
	\label{sec:FormuISPs}
	\begin{figure}
		\centering
		\includegraphics[width = .38\linewidth]{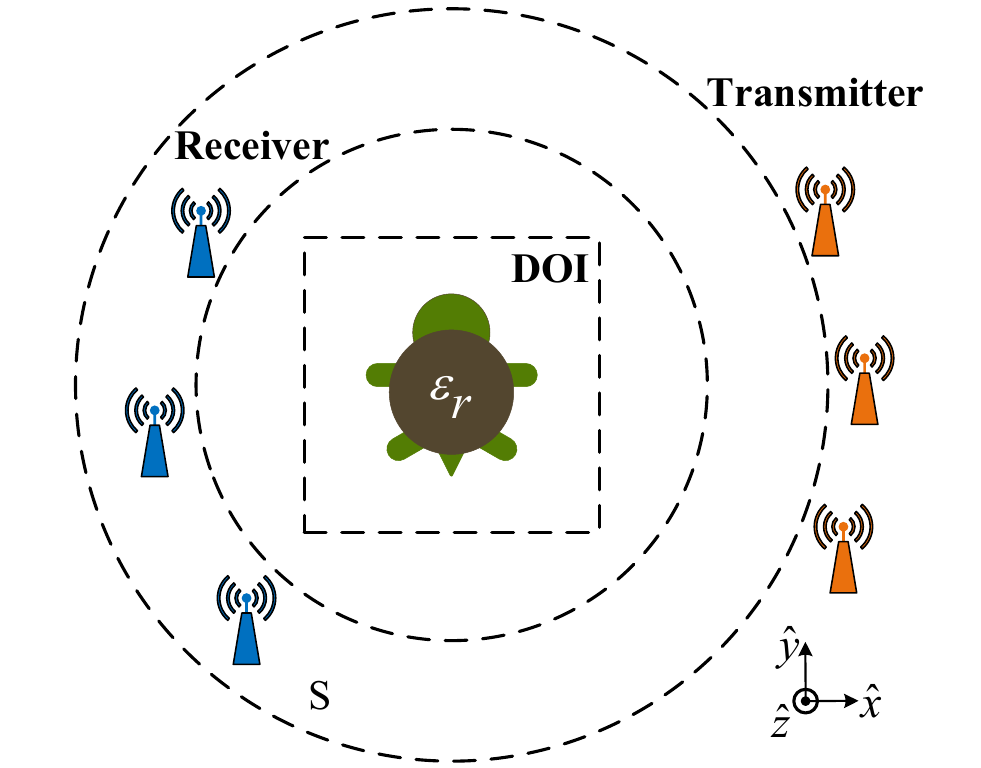}
		\caption{The configuration of 2-D inverse scattering problems.}
		\label{fig:ISPs}
	\end{figure}
	The schematic diagram of the 2-D ISPs is shown in Figure~\ref{fig:ISPs}, where the domain of interest (DOI) is illuminated by transverse magnetic (TM) waves from the ${N_{i}}$ transmitters in turn. Scattered fields are collected by ${N_{s}}$ receivers to reconstruct the electrical properties (e.g., relative permittivity, conductivity, etc.) of the DOI. The problems can be described by state and data equations. The state equation describes the coupling between the incident field and the scatterers, and the multiple scattering effects within the DOI can be expressed as \cite{chen2018computationalEMIS}
	\begin{equation}
		\mathbf{E}^\text{tot}(\mathbf{r}) = \mathbf{E}^\text{inc}(\mathbf{r}) + k_0^2\int_\text{DOI}g(\mathbf{r},\mathbf{r}^\prime)\mathbf{J}(\mathbf{r}^\prime)d\mathbf{r}^\prime, \,\,\mathbf{for} \,\,\mathbf{r}\in\text{DOI},
		\label{eq:stateEqu}
	\end{equation}
	where $\mathbf{E}^\text{tot}(\mathbf{r})$ and $\mathbf{E}^\text{inc}(\mathbf{r})$ are the total and incident electric field at the observation point ${\mathbf{r}}$ within the DOI. $k_0$ denotes the wavenumber in free space. $g$ is the scalar Green's function. $\mathbf{J}$ is the induced current source, given by $\mathbf{J}(\mathbf{r}^\prime)=\boldsymbol{\chi}(\mathbf{r}^\prime)\mathbf{E}^\text{tot}(\mathbf{r}^\prime)$, where $\boldsymbol{\chi}(\mathbf{r}^\prime) = \epsilon_r(\mathbf{r}^\prime)-1$ and $\epsilon_r(\mathbf{r}^\prime)$ is the relative permittivity. The data equation 
	\begin{equation}
		\mathbf{E}^\text{sca}(\mathbf{r}) = k_0^2\int_\text{DOI}g(\mathbf{r},\mathbf{r}^\prime)\mathbf{J}(\mathbf{r}^\prime)d\mathbf{r}^\prime, \,\,\mathbf{for} \,\,\mathbf{r}\in\text{S}
		\label{eq:dataEqu}
	\end{equation} 
	quantifies the scattered fields collected by the receivers. ISPs aim to reconstruct the distribution of $\chi$ from the 
	measured scattered fields and prior physical knowledge of the targets. By discretizing \eqref{eq:stateEqu} and \eqref{eq:dataEqu}, the state and data equations can be rewritten in matrix form as
	\begin{equation}
		\mathbf{E}^\text{tot} = \mathbf{E}^\text{inc} + \mathbf{G}_\text{D}\mathbf{J},
		\label{eq:GreenOperatorStateEqu}
	\end{equation}
	\begin{equation}
		\mathbf{E}^\text{sca} = \mathbf{G}_\text{S}\mathbf{J}.
		\label{eq:GreenOperatorDataEqu}
	\end{equation} 
	where $\mathbf{G}_\text{D}$ and $\mathbf{G}_\text{S}$ denote the discretized integral operators associated with the state and data equations.
	
	\section{Neural Network Solver}
	\label{sec:nnSolver}
	A contrast-source-based physics-driven neural network (CSPDNN) solver is proposed to predict the distribution of the induced current. The network is trained using a composite loss function comprising state consistency, data fidelity, a lower-bound constraint, and total-variation (TV) regularization. Additionally, an adaptive mechanism is designed to dynamically adjust the weight of the TV regularization term.
	
	\subsection{Inversion scheme}
	\label{subsec:nnArch}
	The inversion scheme of the proposed solver is summarized in Algorithm~\ref{alg:CSPDNN}. The network input is formed by concatenating the initial induced current ${\mathbf{J}}_{(0)}$ and the initial relative permittivity ${\boldsymbol{\epsilon}}_{r(0)}$ estimated using the efficient ISP solver backpropagation (BP) method \cite{devaney1982BP}. The network predicts an updated induced current ${\mathbf{J}}_{\boldsymbol{\theta}}$, then the corresponding relative permittivity ${\boldsymbol{\epsilon}}_{r,\boldsymbol{\theta}}$ is subsequently computed. The network parameters are optimized by minimizing loss function. Finally, the predicted permittivity distribution ${\boldsymbol{\epsilon}}_{r}^{\mathrm{pre}}$ is obtained. 
	
	\begin{algorithm}[t]
		\caption{Weight Update of CSPDNN}
		\label{alg:CSPDNN}
		\begin{algorithmic}[1]
			
			\STATE Calculate $\mathbf{G}_\text{D}$ and $\mathbf{G}_\text{S}$;
			\STATE Initialize network parameters $\boldsymbol{\theta} \leftarrow \boldsymbol{\theta}_0$;
			\STATE Construct the network input
			${\mathbf{J}}_{(0)} \oplus {\boldsymbol{\epsilon}}_{r(0)}$;
			
			\STATE Update the equivalent current by CNN:
			${\mathbf{J}}_{\boldsymbol{\theta}}
			=
			\mathcal{F}_{\boldsymbol{\theta}}
			\big({\mathbf{J}}_{(0)} \oplus {\boldsymbol{\epsilon}}_{r(0)}\big)$;
			
			\STATE Compute the relative permittivity distribution:
			${\boldsymbol{\chi}}_{\boldsymbol{\theta}}
			=\frac{\sum\limits_{n=1}^{N_i}\mathbf{{E}}_{\mathbf{tot,}{\boldsymbol{\theta}}}^\mathbf{H}\mathbf{{J}}_{\boldsymbol{\theta}}}{\sum\limits_{n=1}^{N_i}||\mathbf{{E}}^\mathbf{tot}_{{\boldsymbol{\theta}}}||^2}$, ${\boldsymbol{\epsilon}}_{r,\boldsymbol{\theta}}
			=1+\boldsymbol{\chi}_{\boldsymbol{\theta}}$.
			
			\STATE Update $\boldsymbol{\theta}$ by minimizing loss function;
			
			\STATE Output
			${\boldsymbol{\epsilon}}_{r}^{\mathrm{pre}}
			=
			{\boldsymbol{\epsilon}}_{r,\boldsymbol{\theta}}$.
			
		\end{algorithmic}
	\end{algorithm}
	
	\begin{figure}
		\centering
		\includegraphics[width = .5\linewidth]{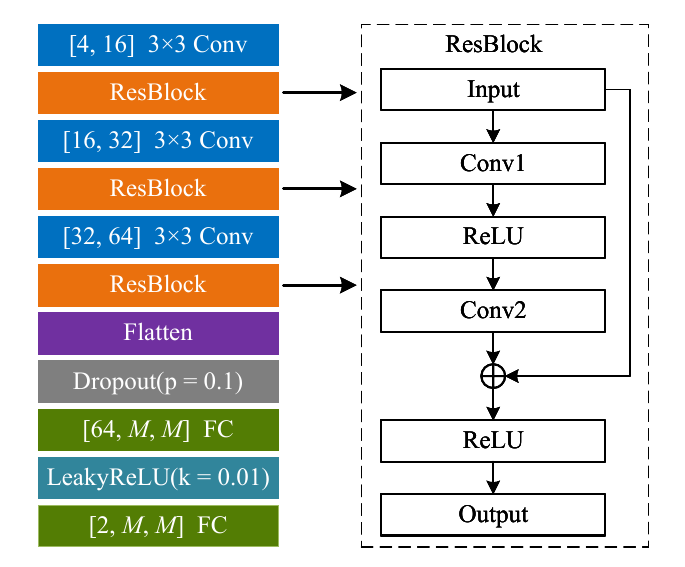}
		\caption{The network architecture of the proposed CSPDNN solver.}
		\label{fig:nnArc}
	\end{figure}
	
	To handle complex-valued data, both the induced current and the permittivity are decomposed into their real and imaginary components, resulting in a four-channel input and a two-channel output. The proposed solver adopts a hybrid convolutional–fully connected architecture that maps an input of size $M \times M$ to an output of the same spatial resolution. As shown in Figure~\ref{fig:nnArc}, the network consists of three convolutional layers with increasing channel numbers of 16, 32, and 64, respectively. Each convolutional layer is followed by a residual block and a LeakyReLU activation to facilitate hierarchical feature extraction. The convolutional features are subsequently flattened and processed by two fully connected layers with dropout and activation functions, enabling global information integration. The batch-size is set to $N_i$. 
	
	\subsection{Loss Function}
	\label{subsec:lossfun}
	The loss function applied by the CSPDNN solver is a summation of four items, \emph{i.e.},
	\begin{equation}
		\text{Loss} = L^{\text{State}}+L^{\text{Data}}+L^{\text{Bound}}+L^{\text{TV}},
		\label{LossFunc}
	\end{equation}
	and the four items are defined as 
	\begin{subequations}
		\begin{align}
			\label{eq:LState}
			&L^\mathrm{State}=\frac{||\mathbf{J}_{\boldsymbol{\theta}} -\boldsymbol\chi_{\boldsymbol{\theta}}(\mathbf{E}^\mathrm{tot}_{\boldsymbol{\theta}})||^2}{||\mathbf{E}^\text{inc}||^2},\\
			\label{eq:LData}
			&L^\mathrm{Data}=\frac{||\mathbf{E}^\text{sca}_{\boldsymbol{\theta}}-\mathbf{E}^\text{sca}_\text{mea}||^2}{||\mathbf{E}^\text{sca}_\text{mea}||^2},\\
			\label{eq:LreEp}
			&L^{\text{Bound}}=\alpha ||\text{ReLU}(1-\mathbf{Re\{\boldsymbol{\epsilon}}_{r,\boldsymbol{\theta}}\})||_1,\\
			\label{eq:LTV}
			&{\color{black}{L^{\text{TV}}=\beta(\Re\{\boldsymbol\chi_{\boldsymbol{\theta}}\})
					f(\Re\{\boldsymbol\chi_{\boldsymbol{\theta}}\})
					+\beta(\Im\{\boldsymbol\chi_{\boldsymbol{\theta}}\})f(\Im\{\boldsymbol\chi_{\boldsymbol{\theta}}\}).}}
		\end{align}
	\end{subequations}
	where $L^\mathrm{State}$ represents the residual of the predicted induced current and the induced current computed from the estimated contrast and total field. $L^{\mathrm{Data}}$ quantifies the discrepancy between the measured scattered fields $\mathbf{E}^\text{sca}_\text{mea}$ and the scattered fields corresponding to the predicted solution $\mathbf{J}_{\boldsymbol{\theta}}$. $L^{\mathrm{Bound}}$ enforces a lower-bound constraint on the real part of the relative permittivity. $L^{\mathrm{TV}}$ is the total-variation regularization term \cite{ma2023inverse}, imposing smoothness on the desired solution, where $f(v)=\sum_{i,j}\sqrt{(v^{i,j+1}-v^{i,j})^2+(v^{i+1,j}-v^{i,j})^2}$, and $\beta(u)=\frac{\beta_0}{M(u)}$, $M(\cdot)$ denotes mean operator. $\alpha$ and $\beta_0$ are hyperparameters set to $1\times10^{-4}$ and $1\times10^{-5}$, which are found to yield stable and satisfactory performance.
	
	\subsection{Training settings}
	\label{subsec:TraSet}
	The solvers are trained by workstation equipped with 128-GB RAM, 3.20-GHz CPU, and NVIDIA GeForce RTX 4090 GPU. The network parameters are optimized using the Adam optimizer with an initial learning rate of $1\times10^{-3}$, which is halved every 1000 epochs. The maximum number of training epochs is set to 1500.
	
	\section{Numerical and Experimental Results}
	\label{sec:Results}
	To evaluate the CSPDNN solver, numerical simulations are conducted on a $0.15\mathrm{m}\times0.15\mathrm{m}$ DOI, which is discretized into a $64\times64$ grid. The imaging system comprises 36 transmitters and 36 receivers uniformly distributed along a circular with a radius of $20\lambda$, sharing the same center as the DOI. The scattered fields are generated using the method of moments (MoM)\cite{Ney1985MoM,Gibson2021MoM} at 4 GHz. 
	
	\subsection{Reconstruction of Complex Scatterers}
	\label{subsec:complexSca}
	
	\begin{figure}[!t]
		\centering
		\includegraphics[width = .8\linewidth]{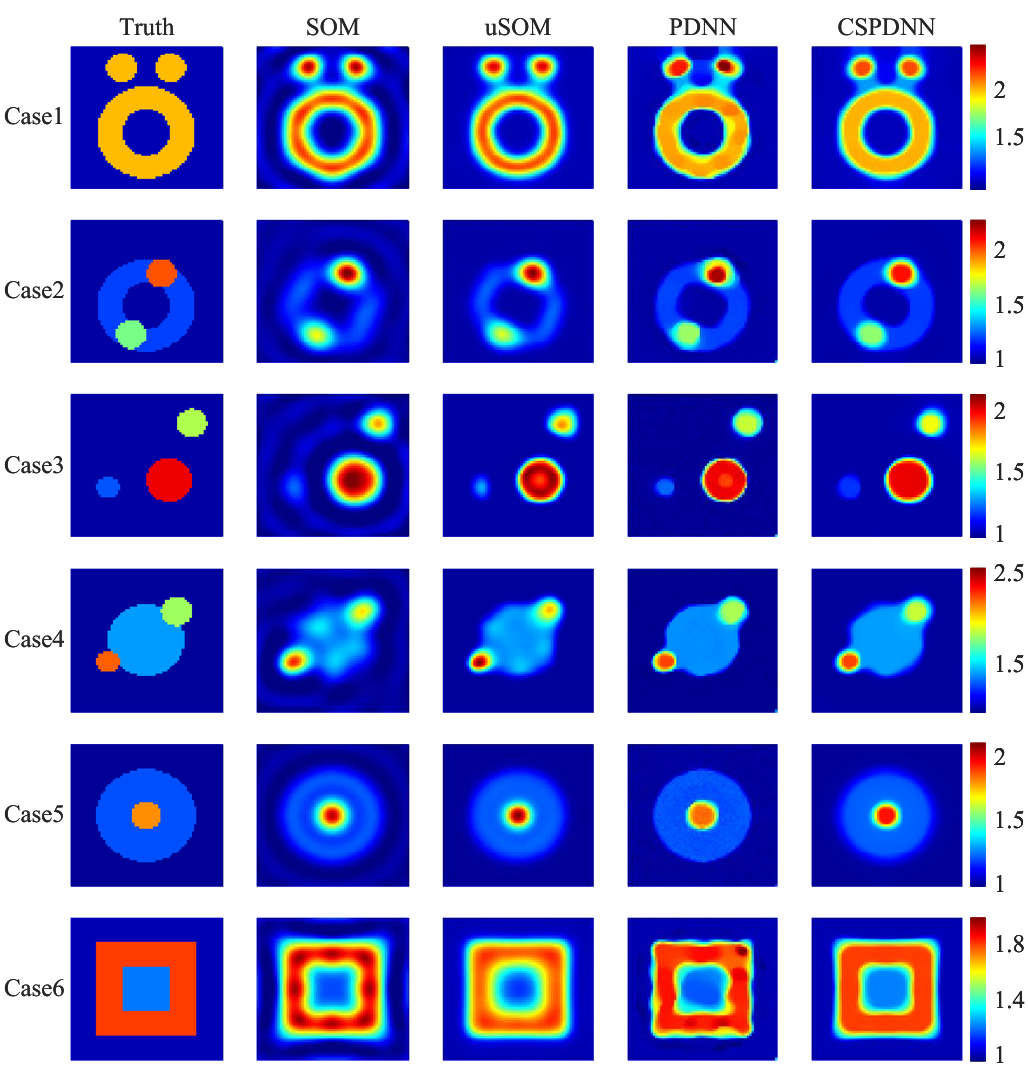}
		\caption{Comparison of imaging results for complex scatterers obtained by SOM, uSOM, PDNN, and CSPDNN solver.}
		\label{fig:Complex}
	\end{figure}
	
	To assess the applicable scope of the proposed CSPDNN solver, six representative cases are considered involving complex scatterer configurations, and the corresponding reconstruction results are shown in Figure~\ref{fig:Complex}.
	
	For Case 1, corresponding to the classical Austria-shaped scatterer, both SOM and uSOM suffer from blurred boundaries, particularly along the ring-shaped structure. PDNN partially recovers the target geometry. However, noticeable shape distortion and locally overestimated contrast values are observed. CSPDNN produces a much sharper boundary with reconstructed values closest to the ground truth, demonstrating superior accuracy. Cases 2 and 4 involve overlapping circular scatterers with different relative permittivities, superimposed on a ring-shaped or circular background. SOM and uSOM are only able to identify the relative spatial location of the scatterers, while the reconstructed shapes are severely distorted. Although both PDNN and CSPDNN successfully recover the overall geometry and size of the scatterers, PDNN exhibits visible local overestimation of contrast in Case 2. CSPDNN achieves more uniform reconstructions with clearly defined boundaries. In Case 3, where scatterers of different sizes and contrast are distributed in the DOI, all solvers can distinguish the locations and contrast differences of the three targets. However, the small low-contrast scatterer on the left is easily misjudged as background artifacts by the results of SOM and uSOM. Both PDNN and CSPDNN yield reconstructions that are much closer to the ground truth, correctly preserving the characteristics of the small low-contrast scatterer. For Case 5, consisting of concentric circular scatterers with different radii and permittivities, SOM and uSOM tend to misjudge the outer weak scatterer as an artifact induced by the stronger inner target. PDNN achieves accurate contrast estimation but produces visibly non-smooth reconstructions. CSPDNN delivers high-quality imaging results with improved smoothness and uniformity, although the reconstructed edges remain slightly less sharp. In Case 6, involving overlapping scatterers with sharp corners, SOM, uSOM, and PDNN are unable to clearly distinguish whether the central response corresponds to a weak scatterer or an artifact induced by a strong scatterer. In contrast, CSPDNN successfully reconstructs the central weak scatterer with a uniform relative permittivity that is clearly separable from the background, yielding the most accurate reconstruction among all the considered solvers. Overall, these results demonstrate that the proposed solver achieves more accurate and physically consistent reconstructions across a wide range of challenging configurations.
	
	\begin{table}[h]
		\centering
		\caption{Comparison of inference time between SOM, uSOM, PDNN and CSPDNN solver}
		\label{tab:compare}
		\begin{tabular}{l|c|c|c|c}
			\toprule
			\textbf{} & \multicolumn{1}{c|}{\textbf{SOM}} & \multicolumn{1}{c|}{\textbf{uSOM}} & \multicolumn{1}{c|}{\textbf{PDNN}} & \multicolumn{1}{c}{\textbf{CSPDNN}} \\  
			\midrule
			Case 1      &98.28s   &78.73s  &115.36s  &27.91s  \\ 
			Case 2      &89.03s   &79.40s  &87.45s   &28.24s  \\
			Case 3      &132.9s   &78.79s  &72.70s   &27.55s  \\
			Case 4      &116.8s   &80.12s  &85.76s   &28.87s  \\
			Case 5      &120.77s  &78.77s  &77.76s   &28.47s  \\
			Case 6      &125.64s  &78.84s  &138.75s  &27.30s  \\
			\bottomrule
		\end{tabular}
	\end{table}
	
	In addition to reconstruction accuracy, the inference time of different solvers is also evaluated. As summarized in Table~\ref{tab:compare}, CSPDNN consistently achieves the shortest inference time across all cases, requiring approximately 28s per reconstruction. Compared with the other solvers, CSPDNN provides an average speedup of approximately $3\times$ to $4\times$. These results demonstrate that CSPDNN significantly improves inference efficiency while maintaining high reconstruction quality.
	
	\subsection{Reconstruction of Lossy Scatterers}
	\label{subsec:lossySca}
	
	\begin{figure}[!t]
		\centering
		\includegraphics[width = .8\linewidth]{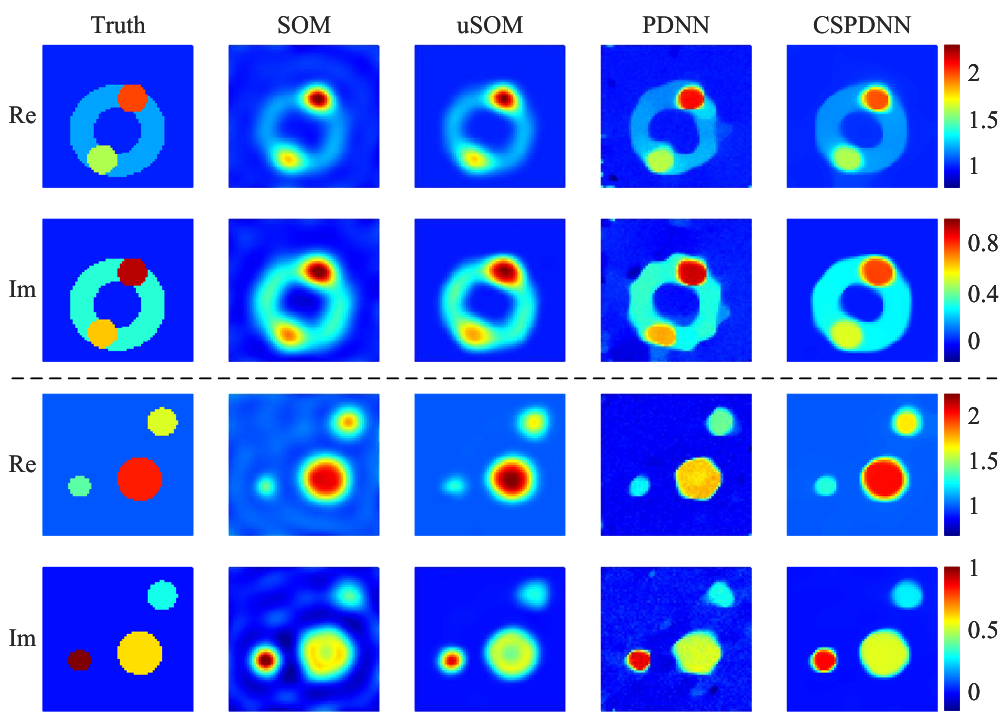}
		\caption{Reconstruction results of lossy scatterers using SOM, uSOM, PDNN, and CSPDNN solver.}
		\label{fig:Lossy}
	\end{figure}
	
	This subsection further investigates more representative lossy scatterers, and two typical reconstruction results are presented in Figure~\ref{fig:Lossy}. As seen, SOM and uSOM exhibit similar reconstruction behaviors. The ring-shaped scatterer in the first case is noticeably distorted by both solvers, although uSOM yields a cleaner background than SOM.
	For the first example, PDNN produces accurate contrast values but suffers from shape distortion and background noise, while in the second case it tends to underestimate the relative permittivity. In contrast, CSPDNN delivers clean backgrounds and accurate reconstructions for both cases, demonstrating the applicability of the proposed solver in lossy scattering scenarios.
	
	\subsection{Influence of Noise Levels}
	\label{subsec:influNoise}
	
	\begin{figure}[!t]
		\centering
		\includegraphics[width = .8\linewidth]{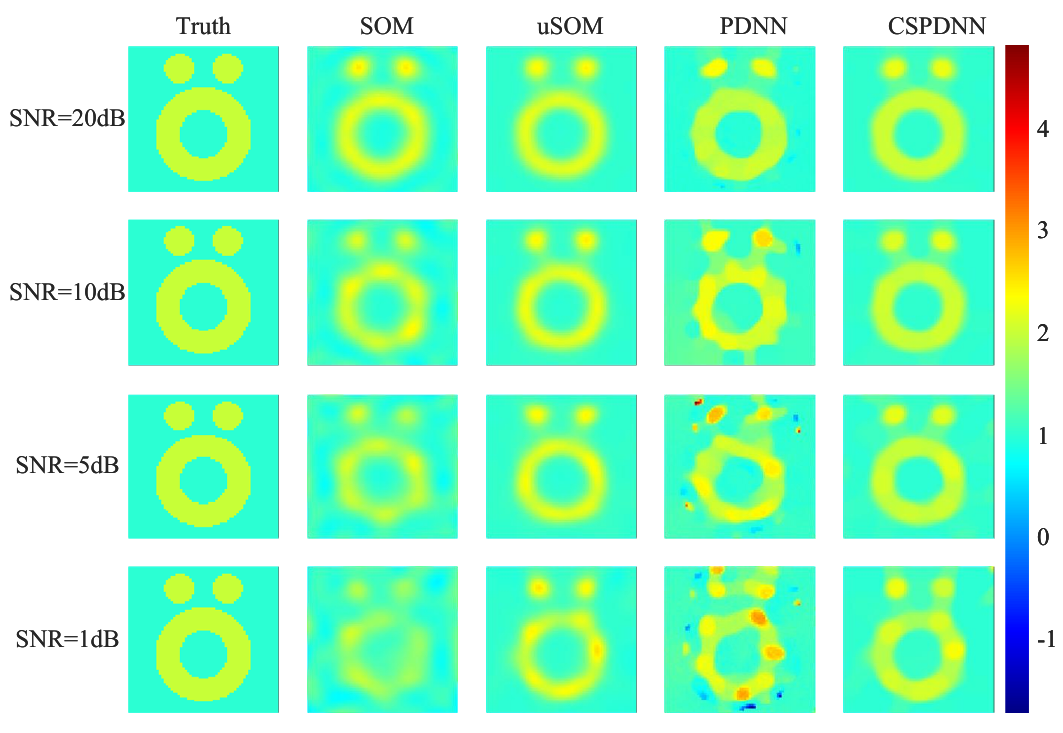}
		\caption{Tests of noise robustness of the ISP solver SOM, uSOM, PDNN and CSPDNN.}
		\label{fig:Noise}
	\end{figure}
	
	Noise robustness is a critical consideration for ISPs. Here, the imaging performance of SOM, uSOM, PDNN, and CSPDNN is evaluated under additive white Gaussian noise with signal-to-noise ratios (SNRs) of 20, 10, 5, and 1 dB. The corresponding reconstruction results are shown in Figure~\ref{fig:Noise}. When SNR = 20 dB, all solvers achieve satisfactory reconstructions. When the SNR decreases to 10 dB, SOM and PDNN exhibit noticeable shape distortions, whereas uSOM and CSPDNN maintain reliable reconstruction quality. Under more severe noise conditions, PDNN suffers from pronounced background artifacts that hinder shape identification. In contrast, uSOM and CSPDNN remain robust even at very low SNR, with CSPDNN providing the most accurate contrast estimation and the clearest boundary definition.
	
	\subsection{Experimental Validation}
	\label{subsec:ExpVal}
	
	\begin{figure}[!t]
		\centering
		\includegraphics[width = .6\linewidth]{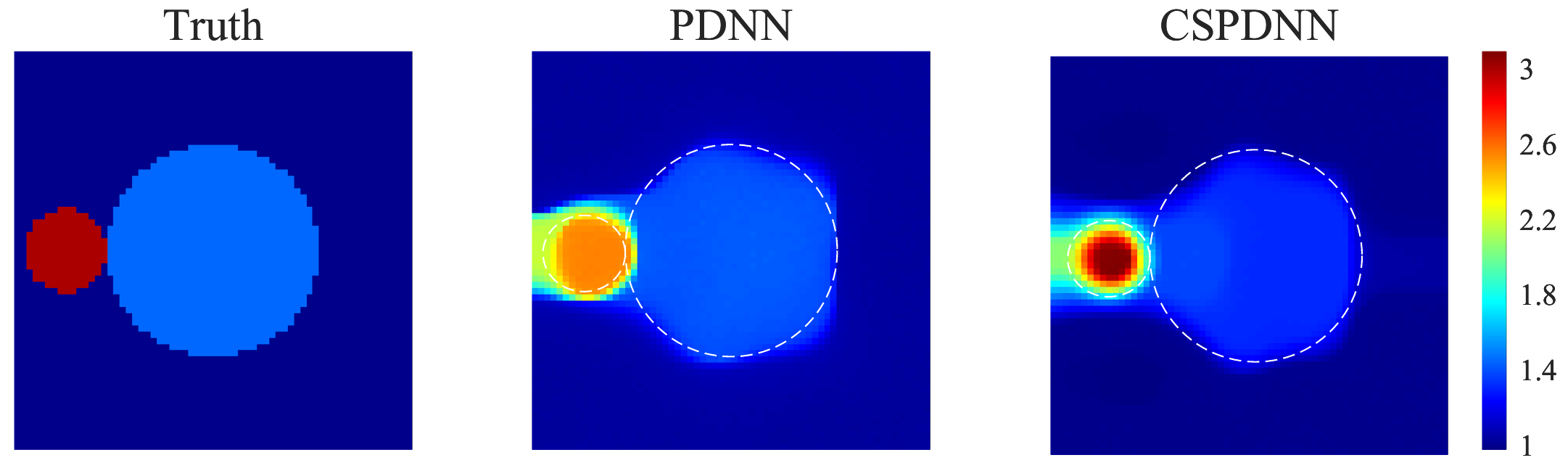}
		\caption{Imaging results based on experimental measurements “FoamDielExt”\cite{Geffrin2005Exp}.}
		\label{fig:exp}
	\end{figure}
	
	The imaging performance of CSPDNN is validated using the experimental dataset “FoamDielExt” provided by the Fresnel Institute \cite{Geffrin2005Exp}. As shown in Figure~\ref{fig:exp}, PDNN exhibits a noticeable underestimation of the strong scatterer on the left side, whereas CSPDNN reconstructs this region more accurately. These results indicate that CSPDNN generalizes well to multiple-scatterer scenarios and high-contrast experimental measurements.
	
	\section{Conclusion}
	\label{sec:conclusion}
	In this paper, a contrast-source-based physics-driven neural network (CSPDNN) is proposed for ISPs. Instead of directly estimating the relative permittivity, the neural network predicts the induced current distribution, thereby bypassing the explicit matrix inversion and significantly improving computational efficiency. Moreover, an adaptive weight updating strategy for the TV regularization term in the loss function is introduced, enabling the solver to automatically adjust to different contrast levels and noise conditions. The imaging performance of CSPDNN is extensively evaluated through representative numerical examples, including lossy and noisy scenarios, and is further validated using experimental measurements. The results demonstrate that CSPDNN achieves accurate, robust, and efficient reconstructions, highlighting its potential for practical applications.
	
	\bibliographystyle{unsrt}
	\bibliography{References}
\end{document}